\documentclass[10pt,letterpaper]{article}
\usepackage{arxiv}
\usepackage[T1]{fontenc}
\usepackage[utf8]{inputenc}
\usepackage{microtype}
\usepackage[dvipsnames]{xcolor}
\usepackage{graphicx,amsmath,amssymb,booktabs,xspace}
\usepackage[numbers,sort&compress]{natbib}
\usepackage[shortlabels,inline]{enumitem}
\usepackage[font=small,labelfont=bf]{caption}
\usepackage[font=footnotesize,skip=3pt,subrefformat=parens]{subcaption}
\usepackage[hyphens]{url}
\usepackage{etoolbox,adjustbox}
\usepackage{tikz}
\usetikzlibrary{arrows.meta,calc,decorations.pathreplacing,positioning}
\newcommand{\bench}{FuGIBench}
\newcommand{\Wait}{\textsc{Wait}}
\newcommand{\Prepare}{\textsc{Prepare}}
\newcommand{\Response}{\textsc{Response}}
\definecolor{fgiwait}{RGB}{218,224,230}
\definecolor{fgiprep}{RGB}{186,216,240}
\definecolor{fgiresp}{RGB}{244,208,156}
\definecolor{fguse}{RGB}{214,228,208}
\definecolor{cvprblue}{rgb}{0.21,0.49,0.74}
\usepackage[breaklinks,colorlinks,allcolors=cvprblue]{hyperref}
\hypersetup{pdftitle={FuGIBench: Evaluating Reliable Preparation in Streaming Video},pdfauthor={Changheng Cai, Zhichao Huang, Song Ren, Zhichen Zhao, Hongsen Liu, Chaoyang Zheng, Wenxi Wu, Yuan Fang, WangWei, Yang Wang, Yuchen Wang}}
\renewcommand{\shorttitle}{FuGIBench}
\renewcommand{\headeright}{Preprint}
\renewcommand{\undertitle}{Preprint}
\title{FuGIBench: Evaluating Reliable Preparation in Streaming Video}
\author{\parbox{\dimexpr\textwidth-2\tabcolsep\relax}{\centering\normalfont
Changheng Cai\textsuperscript{1}, Zhichao Huang\textsuperscript{1}, Song Ren\textsuperscript{1}, Zhichen Zhao\textsuperscript{2},\\[3pt]
Hongsen Liu\textsuperscript{1}, Chaoyang Zheng\textsuperscript{1}, Wenxi Wu\textsuperscript{1}, Yuan Fang\textsuperscript{1},\\[3pt]
WangWei\textsuperscript{3}, Yang Wang\textsuperscript{1}, Yuchen Wang\textsuperscript{1}\\[8pt]
\small\textsuperscript{1}JD.com (jd.com)\\[2pt]
\textsuperscript{2}The Hong Kong University of Science and Technology (ust.hk)\\[2pt]
\textsuperscript{3}Tsinghua University (tsinghua.edu.cn)
}}
\date{}
\BeforeBeginEnvironment{tikzpicture}{\begin{adjustbox}{max width=\linewidth}}
\AfterEndEnvironment{tikzpicture}{\end{adjustbox}}
\begin{document}
\maketitle
\suppressfloats[t]
\begin{abstract}
An assistant watching a worker can prepare for an emerging need before the worker is ready to receive help.
As the need evolves, the assistant must maintain preparation until a response is due, or withdraw it if the need disappears.
Video anticipation and interactive response benchmarks leave this interval between early evidence and response largely unmeasured.
We introduce \bench{} (Future-Grounded Interaction), a streaming-video benchmark in which a model chooses among waiting, reversible preparation, and response given a causal video prefix and a task contract.
The benchmark contains 19,893 synthetic tool-assistance videos across ten tool tasks, in handover, ordinary-activity, and cancelled-need tracks.
One primary score credits sustained preparation, incorporates response-interval alignment, and penalizes activity when the reference prescribes waiting.
On a frozen set of 1,000 episodes, a reference system with a longer preparation lead has poorer response alignment, while similar handover leads accompany different levels of activity during ordinary work in matched scenarios.
Together with cancellation analyses, these results show why evaluating preparation requires tracking decisions as needs develop, become actionable, or disappear.
\end{abstract}

\section{Introduction}
\label{sec:intro}

\input{figures/overview}

While a worker is occupied, an assistant can prepare the tool needed next, maintain that preparation until delivery is due, and withdraw it if the worker changes course.
The same decision arises when preparing the next object or instruction in a physical procedure, a warning before a traffic or pedestrian conflict, or an intervention around an industrial vehicle, hoist, or suspended load.
This paper measures the decision in tool assistance.

Video anticipation predicts forthcoming actions~\cite{Vondrick2016Anticipate,AbuFarha2018When,Damen2018EPIC,Furnari2019RULSTM,Girdhar2021AVT,Grauman2022Ego4D,Assran2025VJEPA2}, and online grounding localizes queried events in an evolving prefix~\cite{Zheng2025HEM}.
Interactive benchmarks judge whether and when a response is warranted~\cite{Shi2026RIVER,Song2026InteractionBench}; proactive systems choose when to respond or delegate~\cite{Yao2026JoyAI}.
World-model benchmarks add generated video, control, and memory~\cite{Fang2026iWorld}.
Future-event accuracy and response timing leave the preceding interval unmeasured: a model may predict correctly early and then drop its preparation, or retain it after the need disappears.

We introduce \bench{} to evaluate preparation grounded in early visual evidence.
Given a causal video prefix and a task contract, a model chooses among \Wait, \Prepare, and \Response, with an identifiable target and response condition.
\bench\ scores whether reversible preparation persists until the response is due and is withdrawn on cancellation.
Figure~\ref{fig:overview} illustrates delivery and cancellation timelines.
The dataset comprises 19,893 synthetic tool-assistance videos across ten tool tasks, with separate positive-handover, ordinary-activity, and cancelled-need tracks.

Preparation lead captures persistence; reliability also depends on response alignment and silence.
We define \emph{preparation lead} as the duration of the uninterrupted \Prepare\ state immediately before the target boundary.
The primary score combines preparation lead with response alignment and inactivity during reference waiting.
Separate diagnostics report preparation, response, and withdrawal behavior.

On the frozen evaluation set, the native video reference has a longer mean preparation lead than the expanded reference ($3.67$ vs.\ $2.08$ seconds), yet its response interval aligns less closely with the reference ($0.083$ vs.\ $0.695$ IoU).
This contrast shows that a longer preparation lead alone does not establish more reliable preparation.
Across 67 workplace-matched pairs, the expanded and native references have similar mean handover leads ($2.27$ and $2.54$ seconds) but are active for $0.15$ and $1.87$ seconds, respectively, during ordinary activity.
Cancellation episodes further reveal whether preparation is withdrawn when the need disappears.
Simple decision policies further separate preparation, response, and silence.

Our contributions are threefold:
\begin{itemize}[leftmargin=*,itemsep=0pt,topsep=0pt,parsep=0pt]
    \item We formulate reliable preparation from early visual evidence as a streaming decision task with explicit target, cancellation, and response semantics.
    \item We instantiate it with 19,893 synthetic tool-assistance videos across ten tool tasks, with annotated stages for handover, ordinary activity, and cancellation.
    \item We evaluate sustained preparation, response alignment, and silence in one primary score, with separate diagnostics for withdrawal.
\end{itemize}

\section{Related Work}
\label{sec:related}

\paragraph{Anticipating future activity.}
Video anticipation predicts a future representation~\cite{Vondrick2016Anticipate}, activity classes and durations~\cite{AbuFarha2018When}, or the next object interaction.
EPIC-KITCHENS supports egocentric action anticipation~\cite{Damen2018EPIC}. Rolling-unrolling LSTMs predict actions at several lead times~\cite{Furnari2019RULSTM}, and Anticipative Video Transformer models temporal progression through causal attention~\cite{Girdhar2021AVT}.
Ego4D forecasts object contact and action sequences~\cite{Grauman2022Ego4D}. V-JEPA~2 learns predictive representations for understanding, anticipation, and planning~\cite{Assran2025VJEPA2}.
These prediction targets establish what may happen next.
\bench\ evaluates the intervening decision: how long preparation persists until a response is due, whether it is withdrawn on cancellation, and which tool is identified.

\paragraph{Streaming perception and interaction.}
Long Short-Term Transformer classifies ongoing activity from a causal history~\cite{Xu2021LSTR}, and Hierarchical Event Memory retains events for online grounding and upcoming-boundary prediction~\cite{Zheng2025HEM}.
VideoLLM-online aligns dialogue with arriving frames~\cite{Chen2024VideoLLM}, while StreamingBench queries understanding at five moments per video~\cite{Lin2024StreamingBench}.
Closer to our decision task, RIVER evaluates retrospective memory, live perception, and proactive response with explicit timing~\cite{Shi2026RIVER}.
InteractionBench scores whether and when a streaming system should speak across queries, triggers, and ongoing updates~\cite{Song2026InteractionBench}, and JoyAI-VL-Interaction chooses silence, response, or delegation~\cite{Yao2026JoyAI}.
Our comparison concerns the scored decisions: response timing evaluates commitment, while \bench\ additionally scores a reversible \Prepare\ state before commitment and its withdrawal when a need disappears.
Tool identity, preparation persistence, response-interval overlap, and cancellation behavior are reported separately.

\paragraph{Physical assistance and procedural tasks.}
Activity anticipation can support reactive robot assistance~\cite{Koppula2013Anticipating}, and HoloAssist records interventions during manipulation~\cite{Wang2023HoloAssist}.
Assembly101 benchmarks anticipation in procedural activities~\cite{Sener2022Assembly101}.
DexH2R studies dynamic human-to-robot handover and grasping~\cite{Wang2025DexH2R}.
\bench\ instead measures a video-based assistant's sustained preparation and withdrawal before tool handover, without evaluating robot execution.

\paragraph{Preparation in other contracts.}
Dashcam models anticipate traffic accidents~\cite{Chan2016Accident}, pedestrian models estimate crossing intentions~\cite{Rasouli2019PIE}, and streaming crime anticipation reasons over developing incidents~\cite{Wang2026Crime}.
Those settings also motivate preparing a warning before it is due and withdrawing it when a conflict clears.
The videos and measurements in this paper instantiate the decision in tool assistance.

\paragraph{World models and controlled comparisons.}
iWorld-Bench evaluates visual generation, trajectory following, and memory through an action-generation interface~\cite{Fang2026iWorld}, whereas \bench\ evaluates decisions from an observed video prefix.
Shortcut learning and causal confusion motivate checking which cues support a prediction~\cite{Geirhos2020Shortcut,DeHaan2019Causal}.
CATER controls scene factors for temporal reasoning~\cite{Girdhar2020CATER}.
Our evaluation rescales generated stage durations to vary event timing, and its workplace-matched comparison holds recorded scene attributes fixed across distinct handover and ordinary-activity videos to contrast preparation for an emerging need with activation during ordinary work.

\section{Reliable Preparation from Streaming Video}
\label{sec:task}

\subsection{Causal Input and Prediction Interface}

An evaluation episode consists of a video $V$ of duration $T$, a task contract $c$, and temporal annotations.
At observation time $t_k$, a model receives only frames with timestamps at or before $t_k$, together with the contract and its allowed history.
The contract specifies the objective, admissible targets, and the condition for a response.
It is fixed before the future portion of the video is observed.
The numeric boundary $\tau$ is a scoring annotation and is not given to the model.
Filenames, generation prompts, full-video summaries, and future annotations are excluded from the model input.

The prediction at each observation is
\begin{equation}
    \hat y_k = (\hat s_k,\hat z_k,\hat \tau_k,\hat e_k),
    \label{eq:prediction}
\end{equation}
where $\hat s_k$ is the decision state, $\hat z_k$ is the target identity, $\hat\tau_k$ is an optional prediction of absolute target time in video seconds, and $\hat e_k$ is an optional event-class prediction.
The three decision states are:
\begin{itemize}[leftmargin=*,itemsep=2pt,topsep=3pt]
    \item \Wait: continue observing without an active preparation or response.
    \item \Prepare: maintain a target-specific, reversible preparation while the observed evidence supports the forthcoming need.
    \item \Response: the model commits to the response prescribed by the contract.
\end{itemize}
The benchmark scores these decisions against the reference stage labels.
\Prepare\ and \Response\ are both called \emph{active} when measuring activity during reference \Wait, whereas only \Prepare\ earns preparation lead.
The held \Response\ interval is evaluated against the annotated response interval.

\subsection{Response Conditions and Event Outcomes}

The contract distinguishes response readiness from the action that follows.
In tool assistance, a worker may prepare a workpiece, receive a tool, and then use it.
The target boundary $\tau$ is the annotated start of handover, when delivery becomes appropriate.
The handover segment defines the reference \Response\ interval, and tool use is a later substage that does not extend it.
The videos, stage boundaries, and measurements in this paper use this tool-assistance contract.
The other contracts described in the introduction use the same split between a response becoming due and the subsequent physical action.
Ordinary and cancelled tool episodes follow \Wait\!$\rightarrow$\!\Wait\ and \Wait\!$\rightarrow$\!\Prepare\!$\rightarrow$\!\Wait.

Each episode belongs to one of three event classes: \emph{target event}, \emph{no event}, or \emph{cancelled event}.
A cancelled event can contain a valid preparation interval followed by evidence that the original need has been abandoned.
It has a cancellation boundary but no realized target boundary for that need.
Unannotated intervals are unknown, and scored episodes require the contract's temporal coverage.

\subsection{State Persistence and Reset}

Decisions are piecewise constant between observations.
An explicit \Wait\ ends preparation; a declared no-update output retains the previous state and target.
Before the first valid decision the state is unknown.
State predictions that remain invalid after the output adapter interrupt reliable preparation, whereas an invalid optional time field is cleared without discarding a valid state and target.

After satisfaction or cancellation, continued preparation is scored against the subsequent reference interval.
Episodes are replayed in order, and model memory is reset between independent episodes.
Adapters declare their silence semantics, history policy, and field parsing.
Section~\ref{sec:experiments} states the observation settings and model interfaces used for the reported comparisons.

\section{Dataset Construction}
\label{sec:dataset}

\subsection{Scenario Design and Seed Curation}

We construct synthetic videos from human-designed scenarios and staged model generation.
Templates specify the workplace, emerging need, response condition, and activity sequence.
The scored annotations are the decoded cuts between generated stages, inherited from the template exemplar.
A seed here is a scenario exemplar, distinct from a run's random seed.
Vision-language checks retain candidate exemplars as \emph{silver seeds}, from which human reviewers select \emph{golden seeds} to choose the templates.
That selection operates at the template level, while the scored boundaries remain the generated stage cuts.
Variants preserve the task contract while varying the workshop description, clothing, and lighting, with three, four, and three options, respectively.
Handover first-frame generation additionally varies the background wall across three options.
The generation pipeline automatically checks each stage and retries it when needed.

\subsection{Dataset and Coverage}

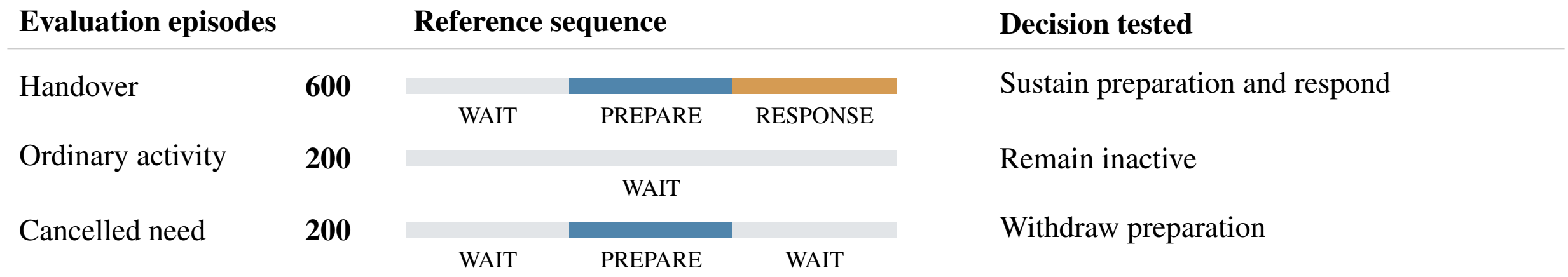
\begin{figure*}[t]
\centering
\begin{tikzpicture}[x=1cm,y=1cm,font=\footnotesize]
\definecolor{panelwait}{HTML}{E2E5E8}
\definecolor{panelprep}{HTML}{5085AC}
\definecolor{panelresp}{HTML}{D59B53}

\node[anchor=west,font=\footnotesize\bfseries] at (0,2.6) {Evaluation episodes};
\node[anchor=west,font=\footnotesize\bfseries] at (4.1,2.6) {Reference sequence};
\node[anchor=west,font=\footnotesize\bfseries] at (10.2,2.6) {Decision tested};
\node[anchor=west] at (0,1.95) {Handover};
\node[anchor=east,font=\footnotesize\bfseries] at (3.7,1.95) {600};
\fill[panelwait] (4.15,1.8699999999999999) rectangle (5.8500000000000005,2.03);
\node[font=\scriptsize,anchor=north] at (5.0,1.8499999999999999) {WAIT};
\fill[panelprep] (5.8500000000000005,1.8699999999999999) rectangle (7.550000000000001,2.03);
\node[font=\scriptsize,anchor=north] at (6.700000000000001,1.8499999999999999) {PREPARE};
\fill[panelresp] (7.550000000000001,1.8699999999999999) rectangle (9.25,2.03);
\node[font=\scriptsize,anchor=north] at (8.4,1.8499999999999999) {RESPONSE};
\node[anchor=west] at (10.2,1.95) {Sustain preparation and respond};
\node[anchor=west] at (0,1.2) {Ordinary activity};
\node[anchor=east,font=\footnotesize\bfseries] at (3.7,1.2) {200};
\fill[panelwait] (4.15,1.1199999999999999) rectangle (5.8500000000000005,1.28);
\fill[panelwait] (5.8500000000000005,1.1199999999999999) rectangle (7.550000000000001,1.28);
\node[font=\scriptsize,anchor=north] at (6.700000000000001,1.0999999999999999) {WAIT};
\fill[panelwait] (7.550000000000001,1.1199999999999999) rectangle (9.25,1.28);
\node[anchor=west] at (10.2,1.2) {Remain inactive};
\node[anchor=west] at (0,0.45) {Cancelled need};
\node[anchor=east,font=\footnotesize\bfseries] at (3.7,0.45) {200};
\fill[panelwait] (4.15,0.37) rectangle (5.8500000000000005,0.53);
\node[font=\scriptsize,anchor=north] at (5.0,0.35) {WAIT};
\fill[panelprep] (5.8500000000000005,0.37) rectangle (7.550000000000001,0.53);
\node[font=\scriptsize,anchor=north] at (6.700000000000001,0.35) {PREPARE};
\fill[panelwait] (7.550000000000001,0.37) rectangle (9.25,0.53);
\node[font=\scriptsize,anchor=north] at (8.4,0.35) {WAIT};
\node[anchor=west] at (10.2,0.45) {Withdraw preparation};
\draw[black!20] (0,2.34) -- (16.2,2.34);
\end{tikzpicture}
\caption{\textbf{Three episode types cover preparation, inactivity, and withdrawal.} The 1,000 evaluation episodes are sampled from the 19,893-video tool collection, with stages rescaled and boundaries remapped. Sequences show stage order, not measured durations.}
\label{fig:dataset}
\end{figure*}

\bench\ contains 19,893 synthetic tool-assistance videos in the tool export inspected on November 6, 2026 (Table~\ref{tab:inventory}).
Figure~\ref{fig:dataset} shows construction and evaluation-set sources.
The ten categories are wrench, tape measure, pliers, tin snips, utility knife, paintbrush, drill, hand saw, hammer, and screwdriver.
The first eight also include ordinary-activity and cancellation tracks.

\begin{table}[t]
\centering
\caption{\textbf{FuGIBench composition.} Tool-assistance videos in the export inspected on November 6, 2026.}
\label{tab:inventory}
\small
\begin{tabular}{@{}lr@{}}
\toprule
Source family & Videos \\
\midrule
Tool handover with initial waiting & 13,194 \\
Tool ordinary activity & 3,502 \\
Tool cancelled need & 3,197 \\
\midrule
\textbf{Total} & \textbf{19,893} \\
\bottomrule
\end{tabular}
\end{table}

The 1,000-episode evaluation set is sampled and frozen before model scoring, comprising 600 handover, 200 ordinary, and 200 cancelled episodes whose source videos are distinct from the 14,002 training videos used by the video references.
Original clips remain grouped with their waiting supplements; positive evaluation videos share scenario titles with the timing-policy source set, from which recorded test-video hashes are excluded.

To vary stage positions, we independently rescale the durations of \Wait, \Prepare, and \Response\ by log-uniform factors in $[0.35,2.5]$, with at least one below $0.8$, one above $1.3$, and a largest-to-smallest ratio of at least $2$.
We rescale an unlabeled tail by a factor within $[0.65,1.4]$.
The same piecewise-linear map transforms every boundary.
Evaluation videos, contracts, boundaries, and scoring code will be released.

\subsection{Tool Assistance Trajectories}

Positive trajectories contain initial waiting, preparation, robot handover, and human tool use.
Adjacent stages meet at a selected frame, defining the scored cuts.
A waiting prefix shifts later timestamps by its recorded offset, keeping the original and extended clip in one episode.
Handover start is the target boundary, and handover and tool use remain distinct substages.

Ordinary activity runs from waiting to waiting while tools remain visible.
Cancellation runs from waiting through preparation back to waiting as the worker switches activity.
Tools and the robot stay in place as the activity changes, giving withdrawal its own boundary.

\section{Evaluation Protocol}
\label{sec:evaluation}

\begin{figure*}[t]
\centering
\begin{tikzpicture}[x=1cm,y=1cm,font=\footnotesize,>=stealth]
\definecolor{panelwait}{HTML}{E2E5E8}
\definecolor{panelprep}{HTML}{5085AC}
\definecolor{panelresp}{HTML}{D59B53}
\node[anchor=east] at (1.85,2.35) {Reference};
\node[anchor=east,align=right] at (1.85,1.45) {Model};
\fill[panelwait] (2,2.23) rectangle (6,2.47);
\fill[panelprep] (6,2.23) rectangle (9,2.47);
\fill[panelresp] (9,2.23) rectangle (15,2.47);
\node[font=\scriptsize] at (4,2.72) {WAIT};
\node[font=\scriptsize] at (7.5,2.72) {PREPARE};
\node[font=\scriptsize] at (12,2.72) {RESPONSE};
\fill[panelwait] (2,1.33) rectangle (3,1.57);
\fill[panelprep] (3,1.33) rectangle (4,1.57);
\fill[panelwait] (4,1.33) rectangle (7,1.57);
\fill[panelprep] (7,1.33) rectangle (9,1.57);
\fill[panelresp] (9,1.33) rectangle (13,1.57);
\fill[panelwait] (13,1.33) rectangle (15,1.57);
\draw[black!65,dashed] (9,.95) -- (9,2.95) node[above] {$\tau$};
\foreach \x in {2,3,4,7,9,13}{\fill[black!60] (\x,1.45) circle (.025);}
\node[font=\scriptsize,text=black!70] at (5.1,1.91) {state held between emissions};
\draw[black!65] (3,1.2) -- (3,1.05) -- (4,1.05) -- (4,1.2);
\node[anchor=north,align=center,font=\scriptsize] at (3.5,.91) {\textbf{Silence $S$}\\Active during reference WAIT};
\draw[<->,black!65] (7,1.05) -- (9,1.05);
\node[anchor=north,align=center,font=\scriptsize] at (8,.91) {\textbf{Preparation $P$}\\Uninterrupted lead $\ell$};
\draw[black!65] (9,3.28) -- (9,3.43) -- (15,3.43) -- (15,3.28);
\node[font=\scriptsize,anchor=south] at (12,3.46) {response union};
\draw[black!65] (9,1.2) -- (9,1.05) -- (13,1.05) -- (13,1.2);
\node[anchor=north,align=center,font=\scriptsize] at (12,.91) {\textbf{Recognition $R$}\\Response overlap: $A=2/3$};
\end{tikzpicture}
\caption{\textbf{Three intervals determine the target-episode score.} Preparation uses the uninterrupted \Prepare\ interval held to $\tau$. Active time in reference \Wait\ lowers silence. Recognition combines event/tool correctness with response-interval IoU: the illustrated overlap covers two-thirds of the union. Dots mark model emissions; each state holds until the next emission or stream end. Tool identity is read at the first active observation before the target. This is a scoring illustration, not a measured result.}
\label{fig:scoring}
\end{figure*}
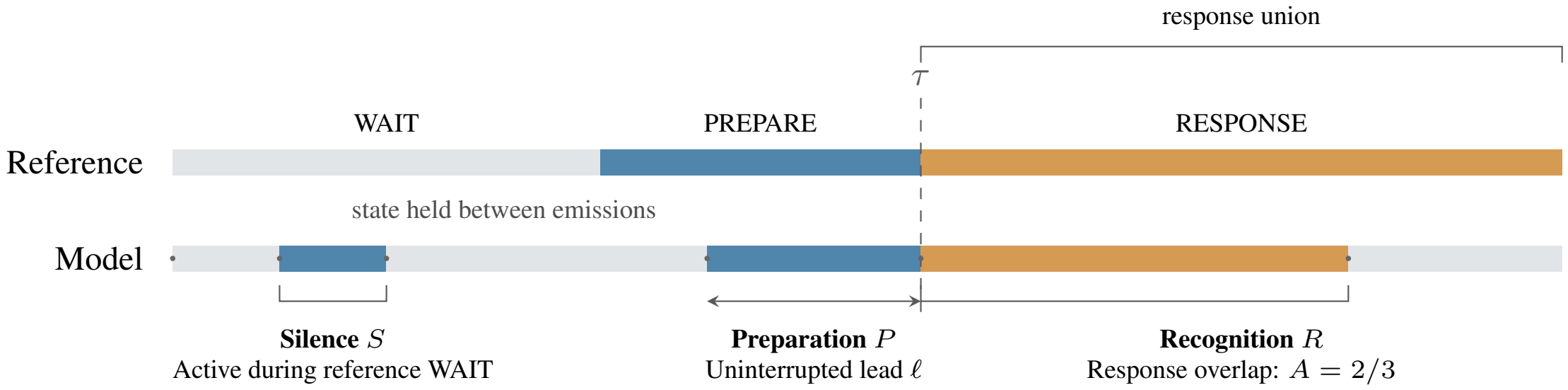

The protocol reports one primary score, the \emph{\bench{} Overall}, from three bounded per-episode components: response-aligned recognition, sustained preparation, and silence when no response is due.
The composition follows $\mathrm{Overall}=(\mathrm{Acc}+H(\mathrm{TA},\mathrm{SC}))/2$ of InteractionBench~\cite{Song2026InteractionBench}, with its behavioral term adapted to preparing before the response.
The episode is the statistical unit.
Figure~\ref{fig:scoring} illustrates the components.

\subsection{Trajectory and Scoring Window}
\label{sec:evaluation:trajectory}

A declared no-update retains the previous decision, and the last decision holds until the reference stream end.
This defines a piecewise-constant \emph{held-state trajectory}.
The reference stages fix the scoring window, and all durations are measured in video seconds, independently of the number of emissions.
The same held behavior receives the same score across emission cadences.

\subsection{Primary Score}
\label{sec:evaluation:components}
\label{sec:evaluation:overall}

Let episode $i$ have reference class $c_i\in\{\text{target},\text{no event},\text{cancelled}\}$.
For a target episode, $\tau_i$ is the response onset, $\rho_i$ its end, and $z_i$ the correct tool.
Let $\hat s_i(t)$ denote the held model state.

\paragraph{Response-aligned recognition $R_i$.}
Content correctness on a target episode is $C_i=\mathbf 1[\hat c_i=\text{target}\ \wedge\ \hat z_i=z_i]$.
Here $\hat z_i$ is the tool named on the earliest active observation strictly before $\tau_i$. An absent tool does not satisfy the tool contract.
This identity check uses that observation, while response alignment evaluates the held \Response\ state.
Let $\mathcal R_i=[\tau_i,\rho_i)$ and let $\hat{\mathcal R}_i$ be the union of all intervals in which the model holds \Response, clipped to the reference stream.
Response alignment is their temporal intersection over union:
\begin{equation}
 A_i=\frac{|\hat{\mathcal R}_i\cap\mathcal R_i|}
 {|\hat{\mathcal R}_i\cup\mathcal R_i|}.
 \label{eq:response-alignment}
\end{equation}
No response gives $A_i=0$.
Recognition is
\begin{equation}
 R_i=\begin{cases}
 C_i A_i, & c_i=\text{target},\\
 \mathbf 1[\hat c_i=c_i], & \text{otherwise.}
 \end{cases}
 \label{eq:recognition}
\end{equation}
Early or overlong response adds time to the union without adding valid overlap; late response misses part of the reference interval.
Response timing enters continuously through $A_i$, without a grace period.
For a reference response of duration $D$, responding $\delta$ seconds early and ending on time gives $A_i=D/(D+\delta)$.

The predicted class $\hat c_i$ is read from the trajectory.
Any \Response\ predicts a target event.
If the model prepares but never responds, the episode is a cancellation when its last recognized state is \Wait, and a target event otherwise.
If it never prepares or responds, the last explicit event class is used, defaulting to no event.
Class accuracy and macro-F1 are separate content diagnostics, whereas $R_i$ also includes response alignment.

\paragraph{Preparation $P_i$.}
Only observations strictly before $\tau_i$ contribute.
Starting from the last such observation, walk backward through the uninterrupted run of \Prepare; let $t_i^{\mathrm{prep}}$ be its first timestamp.
Then
\begin{equation}
 \ell_i=\begin{cases}
 \tau_i-t_i^{\mathrm{prep}}, & \text{the trailing run exists},\\
 0, & \text{otherwise.}
 \end{cases}
 \label{eq:lead}
\end{equation}
A \Wait, \Response, or invalid state interrupts the run.
The lead measures state persistence, including any part of the run in reference \Wait, while tool identity is evaluated separately in $C_i$.
A supplementary joint diagnostic requires the correct tool throughout the trailing \Prepare\ run.
The bounded preparation component is
\begin{equation}
 P_i=\min\!\big(1,\ \ell_i/\Lambda\big),\qquad \Lambda=10\,\mathrm{s}.
 \label{eq:preparation}
\end{equation}
One global scale uses video-time seconds across episodes.
The reference preparation interval has median $4.25$\,s and 90th percentile $10.5$\,s on the 600 targets.
The response factor is continuous in interval overlap, whereas preparation has the distinct requirement of remaining in \Prepare\ at the last pre-target observation.

\paragraph{Silence $S_i$.}
Both \Prepare\ and \Response\ count as active when the reference prescribes \Wait.
With $f_i$ the held active seconds inside reference \Wait\ intervals and $w_i$ the reference \Wait\ duration,
\begin{equation}
 S_i=\max\!\big(0,\ 1-f_i/w_i\big),
 \label{eq:safety}
\end{equation}
and $S_i=1$ when $w_i=0$.
A cancelled episode's valid preparation interval is not penalized, while activity continued into the subsequent \Wait\ lowers $S_i$.

\paragraph{Overall.}

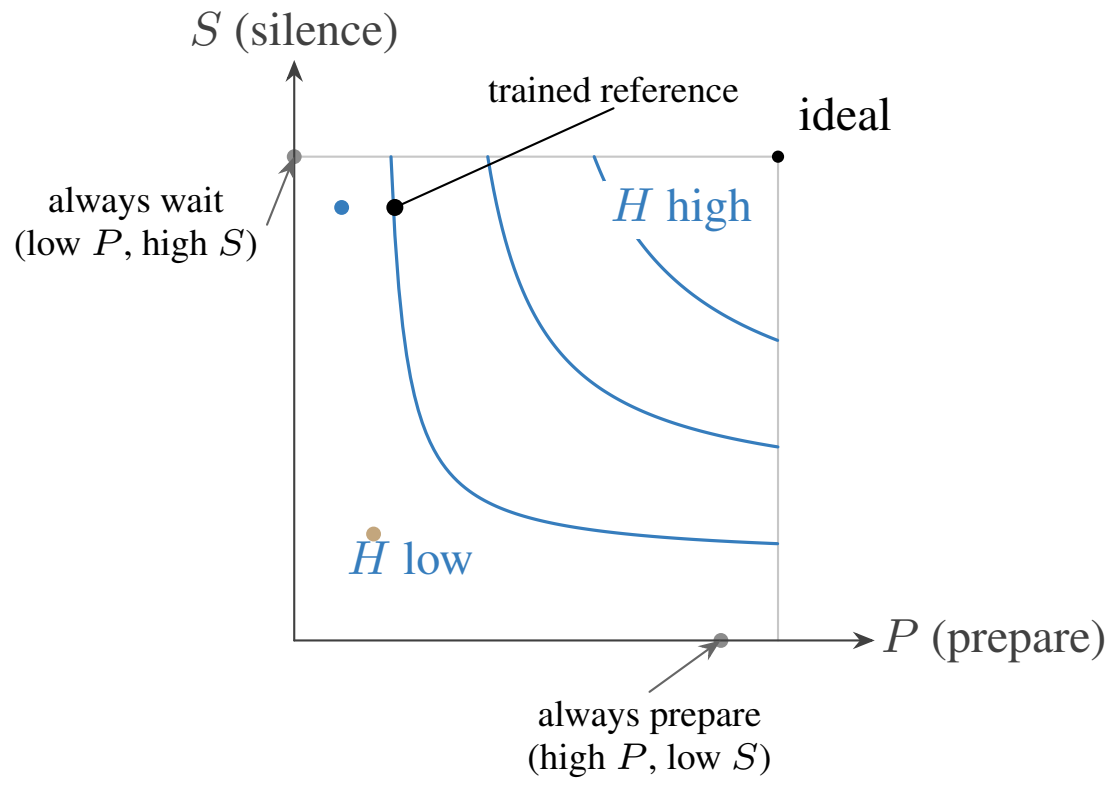
\begin{figure}[t]
\centering
\begin{tikzpicture}[font=\footnotesize,line cap=round,line join=round,
  >={Stealth[length=1.45mm,width=1.15mm]},
  callout/.style={align=center,font=\scriptsize,inner sep=1.5pt}]
  \begin{scope}[shift={(2.20,1.20)},x=3.20cm,y=3.20cm]
    \draw[black!22] (1,0) -- (1,1) -- (0,1);
    % H(P,S) is constant along S = aP / ((1+a)P-a), where a = S at P=1.
    \foreach \a in {0.20,0.40,0.62} {
      \draw[cvprblue,semithick,domain=\a:1,samples=100]
        plot (\x,{\a*\x/((1+\a)*\x-\a)});
    }
    \node[fill=white,inner sep=1.4pt,text=cvprblue] at (0.24,0.17) {$H$ low};
    \node[fill=white,inner sep=1.4pt,text=cvprblue] at (0.80,0.89) {$H$ high};
    % Representative measured operating points; full model values are in Table 3.
    \fill[black] (0.208,0.895) circle (1.6pt);
    \draw[black,thin] (0.208,0.895) -- (0.66,1.10) node[anchor=south,inner sep=1pt,font=\scriptsize] {trained reference};
    \fill[cvprblue] (0.098,0.895) circle (1.4pt);
    \fill[fgiresp!80!black] (0.164,0.220) circle (1.4pt);
    \fill[black!45] (0,1) circle (1.4pt);
    \fill[black!45] (0.882,0) circle (1.4pt);
    \fill (1,1) circle (1.15pt);
    \draw[->,black!75] (0,0) -- (1.20,0) node[anchor=west,inner sep=1pt] {$P$ (prepare)};
    \draw[->,black!75] (0,0) -- (0,1.20) node[anchor=south,inner sep=1pt] {$S$ (silence)};
    \node[anchor=south west,inner sep=1pt] at (1.03,1.04) {ideal};
  \end{scope}
  % degenerate policies sit outside the axes, with a short leader into each corner
  \node[callout,anchor=east] (aw) at (2.02,3.95) {always wait\\(low $P$, high $S$)};
  \draw[->,thin,black!60] (aw.east) -- (2.20,4.40);
  \node[callout,anchor=north] (ap) at (4.55,0.86) {always prepare\\(high $P$, low $S$)};
  \draw[->,thin,black!60] (ap.north) -- (5.02,1.20);
\end{tikzpicture}
\caption{\textbf{Preparation and silence operating points.} Markers show mean $P$ over 600 targets and mean $S$ over all 1,000 episodes for the trained reference, Qwen2.5-Omni (\textcolor{cvprblue}{$\bullet$}), Gemma-3n (\textcolor{fgiresp!80!black}{$\bullet$}), and two constant policies. Contours illustrate the harmonic-mean geometry for individual target episodes. Evaluating $H$ at a plotted aggregate point does not recover mean behavioral reward, which is computed per episode.}
\label{fig:operating}
\end{figure}

With $H(a,b)=2ab/(a+b)$ and $H(0,\cdot)=0$, the per-episode score is
\begin{equation}
 O_i=\begin{cases}
 (R_i+H(P_i,S_i))/2, & c_i=\text{target},\\[2pt]
 (R_i+S_i)/2, & \text{otherwise,}
 \end{cases}
 \label{eq:overall}
\end{equation}
and $\text{\bench{} Overall}=\frac{1}{N}\sum_i O_i$.
On target episodes, the harmonic mean requires both preparation and silence for a positive behavioral reward (Fig.~\ref{fig:operating}).
Evaluating $H$ at the plotted component means does not recover the per-episode behavioral reward.
Overall combines sustained preparation lead with inactivity during reference \Wait.
Preparation extending into reference \Wait\ can increase lead while lowering $S_i$ through the false-activation cost.
The reference trajectory defines the intervals for these measurements, while the model trajectory describes the decisions being evaluated.
Together with response alignment, the score evaluates preparation as a streaming decision across these attributes. We report $P$ and $S$ separately to show their trade-off.
For target episodes, recognition combines response-interval alignment with event-class correctness and the correctness of the tool named at the first active observation before the target.
Non-target episodes have no preparation obligation, so their behavioral term is silence alone.
Recognition and behavior have equal weight, and components combine per episode before averaging, without fitted mixture weights.

\subsection{Diagnostic Measures}
\label{sec:evaluation:diagnostics}

We report $R$ and $S$ over all episodes, and $P$, mean lead, and lead coverage over all target episodes, including zero leads. Lead coverage is the fraction of target episodes with positive lead.
Preparation, response, and withdrawal are reported separately in the experiments and supplementary tables.
Response diagnostics comprise mean $A_i$, the fraction with positive response overlap, premature commitment, missing response, and error in the first response onset relative to $\tau_i$.
Premature commitment means the first \Response\ occurs before $\tau_i$. Response time outside the reference interval adds to the union without adding overlap.
Withdrawal diagnostics comprise trajectory-derived cancellation-class recall and mean active seconds in the reference \Wait\ following cancellation, over all cancelled episodes.
We additionally report how often a model prepares before cancellation. Among episodes in which the model still holds \Prepare\ immediately before cancellation, we report how often it subsequently returns to \Wait, how often it remains in \Wait\ through stream end, and its first-return delay.
Delay uses observed returns; episodes without a return remain in the return-rate denominator.
False activation is additionally reported over all scorable \Wait\ time.
For optional absolute target-time predictions, TTE is the absolute error of the first such prediction made while active strictly before $\tau_i$, reported with prediction coverage; absent predictions are N/A.

Supplementary examples contrast sustained preparation with early or missing response.
All reported results use \texttt{fgi.metrics.episode.v4} on the same frozen 1,000-episode tool-assistance set.
Each run records polling or native emission mode and frame period.

\section{Experiments and Diagnostic Analysis}
\label{sec:experiments}

\subsection{Experimental Setup}

We evaluate preparation, response alignment, and withdrawal on 600 handovers, 200 ordinary activities, and 200 cancellations.
\begin{figure}[t]
\centering
\begin{tikzpicture}[x=1.18cm,y=.70cm,font=\footnotesize,
  bar/.style={line width=4pt,line cap=butt}]
\definecolor{modelexpanded}{HTML}{397E78}
\definecolor{modelnative}{HTML}{777185}
  \node[anchor=west,font=\footnotesize\bfseries] at (0,4.55) {Handover: preparation lead};
  \node[anchor=west,font=\footnotesize\bfseries] at (0,1.95) {Ordinary activity: active time};
  \foreach \x in {0,1,2,3} {
    \draw[black!10] (\x,2.66) -- (\x,4.15);
    \draw[black!10] (\x,.05) -- (\x,1.55);
    \node[font=\scriptsize,anchor=north] at (\x,2.64) {\x};
    \node[font=\scriptsize,anchor=north] at (\x,.03) {\x};
  }
  \draw[bar,modelexpanded] (0,3.90) -- (2.27,3.90) node[anchor=west,black,font=\scriptsize] {2.27};
  \draw[bar,modelnative] (0,3.15) -- (2.54,3.15) node[anchor=west,black,font=\scriptsize] {2.54};
  \draw[bar,modelexpanded] (0,1.30) -- (.15,1.30) node[anchor=west,black,font=\scriptsize] {0.15};
  \draw[bar,modelnative] (0,.55) -- (1.87,.55) node[anchor=west,black,font=\scriptsize] {1.87};
  \node[font=\scriptsize,anchor=west] at (3.1,2.54) {seconds};
  \node[font=\scriptsize,anchor=west] at (3.1,-.07) {seconds};
  \draw[bar,modelexpanded] (0,-.70) -- (.25,-.70);
  \node[anchor=west,font=\scriptsize] at (.32,-.70) {Expanded video};
  \draw[bar,modelnative] (1.85,-.70) -- (2.10,-.70);
  \node[anchor=west,font=\scriptsize] at (2.17,-.70) {Native video + LoRA};
\end{tikzpicture}
\caption{\textbf{Preparation and activation in matched scenarios.} Mean preparation lead on handovers and mean active time on ordinary activity are shown separately for the 67 pairs with matched recorded workplace attributes. Both references receive video. Full descriptive results are supplementary.}
\label{fig:matched}
\end{figure}
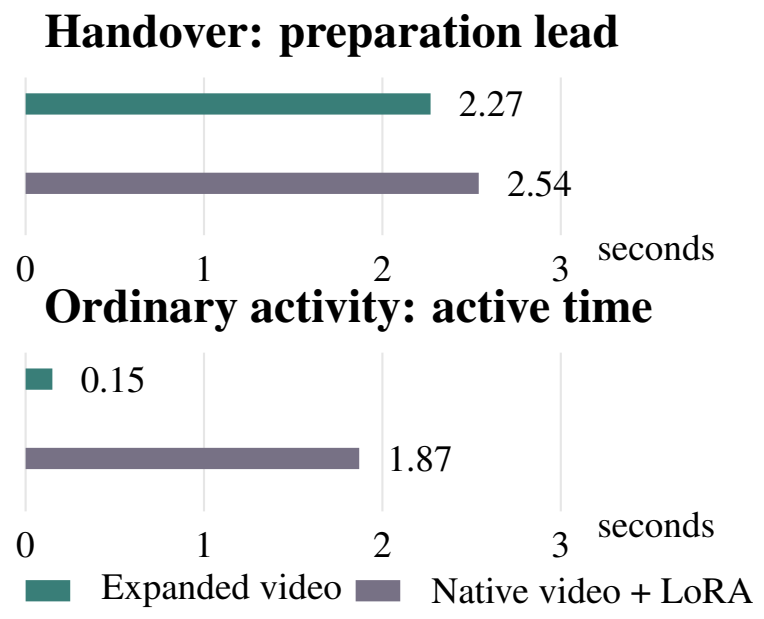

All learned systems receive video observations without audio.
Table~\ref{tab:main} separates models without benchmark-specific training from trained video references.
Frames arrive every second, except VideoChat3 at $0.5$\,s. The session resets at each episode.
Task instructions request a state and tool from general-purpose models, while native streaming systems retain their emission and no-update mechanisms.

\begin{table*}[t]
\centering
\caption{\textbf{Tool-assistance results on 1,000 episodes} (600 handovers, 200 ordinary activities, 200 cancellations). $R$, $P$, and $S$ are response-aligned recognition, preparation, and silence. On targets, $R$ combines event-class and tool correctness with response-interval overlap. The tool is named at the first active observation before the target. Lead and coverage use all targets. All learned systems receive video. Trained references use a fixed tool-vocabulary mapping. Adapted denotes an output adapter without ground truth; reference configurations are in Section~\ref{sec:experiments}.}
\label{tab:main}
\small
\begin{tabular}{@{}llrrrrrr@{}}
\toprule
Model & Mode & Overall $\uparrow$ & $R$ $\uparrow$ & $P$ $\uparrow$ & $S$ $\uparrow$ & Lead $\uparrow$ & Cov. $\uparrow$ \\
 & & $[0,1]$ & $[0,1]$ & $[0,1]$ & $[0,1]$ & seconds & fraction \\
\midrule
\multicolumn{8}{@{}l}{\textit{Evaluated without benchmark-specific training (zero-shot / prompt-adapted)}}\\
Qwen3-VL-8B-Instruct~\cite{Bai2025Qwen3VL} & polled & 0.298 & 0.213 & 0.004 & 0.916 & 0.04 & 0.06 \\
Qwen2.5-Omni-7B~\cite{Xu2025QwenOmni} & polled & 0.301 & 0.205 & 0.098 & 0.895 & 1.19 & 0.12 \\
LLaVA-OneVision-2-8B-Instruct~\cite{An2026LlavaOV2} & polled & 0.300 & 0.200 & 0.005 & 0.995 & 0.05 & 0.01 \\
JoyAI-VL-Interaction~\cite{Yao2026JoyAI} & native & 0.300 & 0.200 & 0.000 & 0.999 & 0.00 & 0.00 \\
MMDuet2~\cite{Wang2025MMDuet2} & native & 0.299 & 0.200 & 0.000 & 0.999 & 0.00 & 0.00 \\
LiveCC-7B-Instruct~\cite{Chen2025LiveCC} & native, adapted & 0.217 & 0.036 & 0.000 & 0.998 & 0.00 & 0.00 \\
VideoChat3-4B~\cite{Li2026VideoChat3} & native & 0.200 & 0.000 & 0.000 & 1.000 & 0.00 & 0.00 \\
MOSS-Video~\cite{Wang2026MOSSVideo} & native, adapted & 0.128 & 0.023 & 0.051 & 0.677 & 0.52 & 0.20 \\
Gemma-3n-E4B-it~\cite{Google2025Gemma3n} & polled & 0.092 & 0.053 & 0.164 & 0.220 & 1.89 & 0.34 \\
MiniCPM-o-2\_6~\cite{OpenBMB2025MiniCPMO} & polled, adapted & 0.091 & 0.013 & 0.107 & 0.368 & 1.16 & 0.25 \\
\midrule
\multicolumn{8}{@{}l}{\textit{References with task-specific training}}\\
Trained reference & polled & 0.645 & 0.741 & 0.208 & 0.895 & 2.08 & 0.56 \\
JoyAI + task head~\cite{Yao2026JoyAI} & polled & 0.363 & 0.324 & 0.514 & 0.577 & 5.36 & 0.93 \\
\midrule
\multicolumn{8}{@{}l}{\textit{Short-schedule video references}}\\
Expanded video reference & polled & 0.622 & 0.710 & 0.208 & 0.880 & 2.08 & 0.56 \\
Native video + LoRA & polled & 0.218 & 0.045 & 0.350 & 0.661 & 3.67 & 0.68 \\
\bottomrule
\end{tabular}
\end{table*}

JoyAI + task head is a task-trained structured prediction head over frozen JoyAI visual and task-query features.
The native video reference fine-tunes JoyAI's decoder LoRA while retaining its visual path.
The expanded reference adds frozen V-JEPA~2 features through a learned bridge.
Both read eight recent frames at $1$\,fps, while the expanded reference also receives up to eight older four-frame chunks.
Their training set is a subset of the tool-video collection, comprising handover, ordinary-activity, and cancellation videos.
They share 14,002 training videos and 547 joint updates with the same sampling schedule and seed.
The expanded reference first completes bridge alignment and then trains bridge and decoder LoRA, whereas the native reference starts directly with LoRA.
Their visual encoders stay frozen, and the comparison concerns the resulting systems.
The separate trained reference uses the expanded architecture with 2,688 joint updates.
The supplement details task instructions, history, decoding, and output adapters.

\subsection{Main Results and Behavioral Analysis}
Across the ten models without benchmark-specific training, Overall spans $0.091$--$0.301$.
Several models score close to the always-wait policy's $0.300$, but exhibit different behaviors: Qwen2.5-Omni scores $0.301$, with a $1.19$\,s mean lead, $0.253$ response alignment, and $0.895$ silence, whereas always-wait has zero lead, zero response alignment, and silence of $1$.
Qwen3-VL-8B-Instruct reaches $0.298$, with silence $0.916$, a $0.04$\,s mean lead, and response alignment $0.239$.
Gemma-3n-E4B-it reaches $0.092$, with silence $0.220$, a $1.89$\,s mean lead, and alignment $0.253$.

The video references illustrate why preparation lead must be read together with response alignment.
The native video reference has a longer mean lead than the expanded reference ($3.67$ vs.\ $2.08$\,s), while their response alignments are $0.083$ and $0.695$, respectively.
Their Overall scores are $0.218$ and $0.622$, as the score accounts for response alignment and inactivity during reference waiting alongside lead.
JoyAI + task head combines a $5.36$\,s mean lead and $0.93$ coverage with $0.536$ response alignment and $0.577$ silence, illustrating a different combination of preparation, response, and waiting behavior.

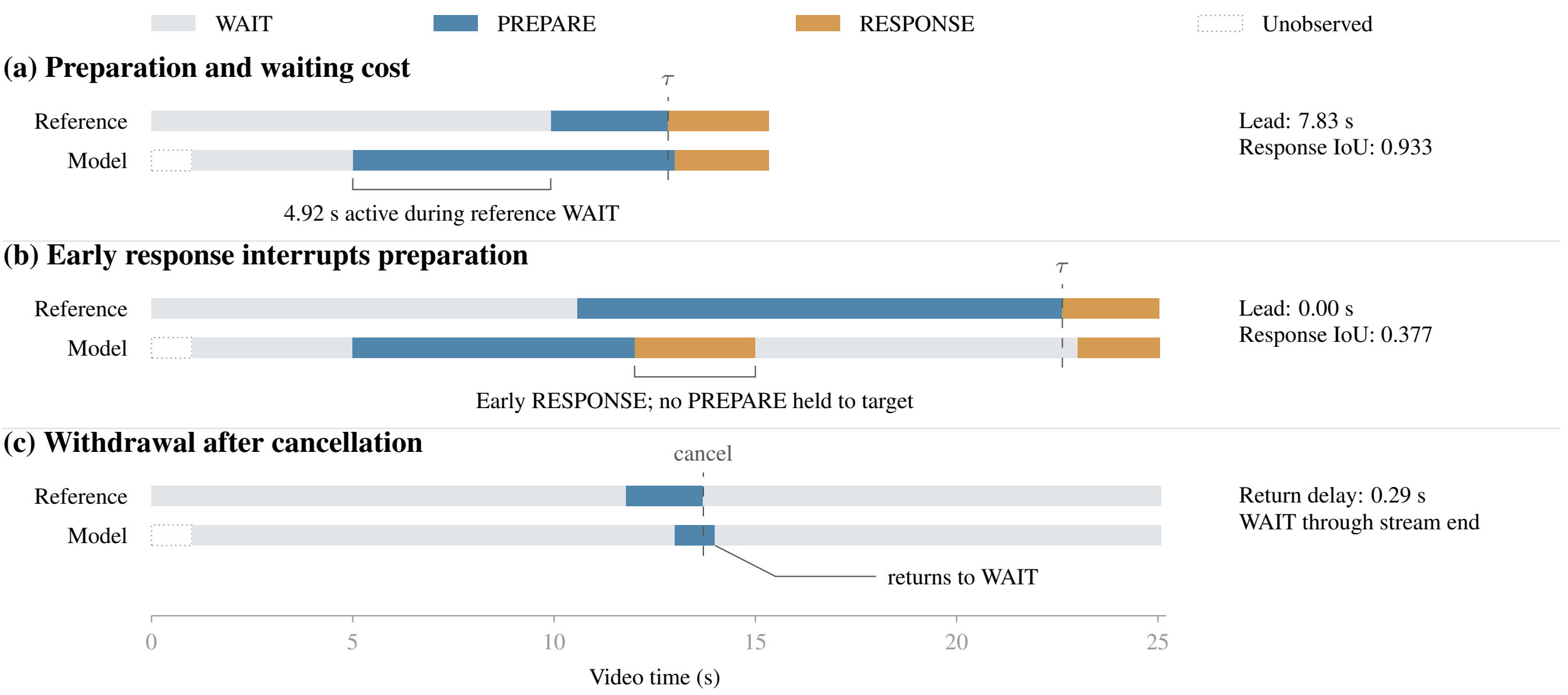
\begin{figure*}[t]
\centering
\begin{tikzpicture}[x=0.43cm,y=1cm,font=\small,>=stealth]
\definecolor{traceblue}{HTML}{5085AC}
\definecolor{traceorange}{HTML}{D59B53}
\definecolor{tracewait}{HTML}{E2E5E8}
\fill[tracewait] (0,6.0) rectangle (1.1,6.17);
\node[anchor=west,font=\scriptsize] at (1.3,6.085) {WAIT};
\fill[traceblue] (7,6.0) rectangle (8.1,6.17);
\node[anchor=west,font=\scriptsize] at (8.3,6.085) {PREPARE};
\fill[traceorange] (16,6.0) rectangle (17.1,6.17);
\node[anchor=west,font=\scriptsize] at (17.3,6.085) {RESPONSE};
\fill[white] (26,6.0) rectangle (27.1,6.17);
\draw[black!40,densely dotted] (26,6.0) rectangle (27.1,6.17);
\node[anchor=west,font=\scriptsize] at (27.3,6.085) {Unobserved};
\node[anchor=west,font=\small\bfseries,inner sep=0pt] at (-3.7,5.590) {(a) Preparation and waiting cost};
\node[anchor=east,font=\scriptsize] at (-.3,5.040) {Reference};
\fill[tracewait] (0.000000,4.930) rectangle (9.916667,5.150);
\fill[traceblue] (9.916667,4.930) rectangle (12.833333,5.150);
\fill[traceorange] (12.833333,4.930) rectangle (15.333333,5.150);
\node[anchor=east,font=\scriptsize] at (-.3,4.620) {Model};
\fill[white] (0.000000,4.510) rectangle (1.000000,4.730);
\draw[black!40,densely dotted] (0.000000,4.510) rectangle (1.000000,4.730);
\fill[tracewait] (1.000000,4.510) rectangle (5.000000,4.730);
\fill[traceblue] (5.000000,4.510) rectangle (13.000000,4.730);
\fill[traceorange] (13.000000,4.510) rectangle (15.333333,4.730);
\draw[black!65,dashed] (12.833333,4.410) -- (12.833333,5.300) node[above,font=\scriptsize] {$\tau$};
\draw[black!65] (5.000000,4.430) -- (5.000000,4.310) -- (9.916667,4.310) -- (9.916667,4.430);
\node[anchor=north,font=\scriptsize] at (7.458333,4.270) {4.92 s active during reference WAIT};
\node[anchor=north west,align=left,font=\scriptsize,inner sep=0pt] at (27,5.140) {Lead: 7.83 s\\Response IoU: 0.933};
\draw[black!12] (-3.7,3.760) -- (35,3.760);
\node[anchor=west,font=\small\bfseries,inner sep=0pt] at (-3.7,3.590) {(b) Early response interrupts preparation};
\node[anchor=east,font=\scriptsize] at (-.3,3.040) {Reference};
\fill[tracewait] (0.000000,2.930) rectangle (10.583333,3.150);
\fill[traceblue] (10.583333,2.930) rectangle (22.625000,3.150);
\fill[traceorange] (22.625000,2.930) rectangle (25.041667,3.150);
\node[anchor=east,font=\scriptsize] at (-.3,2.620) {Model};
\fill[white] (0.000000,2.510) rectangle (1.000000,2.730);
\draw[black!40,densely dotted] (0.000000,2.510) rectangle (1.000000,2.730);
\fill[tracewait] (1.000000,2.510) rectangle (5.000000,2.730);
\fill[traceblue] (5.000000,2.510) rectangle (12.000000,2.730);
\fill[traceorange] (12.000000,2.510) rectangle (15.000000,2.730);
\fill[tracewait] (15.000000,2.510) rectangle (23.000000,2.730);
\fill[traceorange] (23.000000,2.510) rectangle (25.041667,2.730);
\draw[black!65,dashed] (22.625000,2.410) -- (22.625000,3.300) node[above,font=\scriptsize] {$\tau$};
\draw[black!65] (12.000000,2.430) -- (12.000000,2.310) -- (15.000000,2.310) -- (15.000000,2.430);
\node[anchor=north,font=\scriptsize] at (13.500000,2.270) {Early RESPONSE; no PREPARE held to target};
\node[anchor=north west,align=left,font=\scriptsize,inner sep=0pt] at (27,3.140) {Lead: 0.00 s\\Response IoU: 0.377};
\draw[black!12] (-3.7,1.760) -- (35,1.760);
\node[anchor=west,font=\small\bfseries,inner sep=0pt] at (-3.7,1.590) {(c) Withdrawal after cancellation};
\node[anchor=east,font=\scriptsize] at (-.3,1.040) {Reference};
\fill[tracewait] (0.000000,0.930) rectangle (11.791667,1.150);
\fill[traceblue] (11.791667,0.930) rectangle (13.708333,1.150);
\fill[tracewait] (13.708333,0.930) rectangle (25.083333,1.150);
\node[anchor=east,font=\scriptsize] at (-.3,0.620) {Model};
\fill[white] (0.000000,0.510) rectangle (1.000000,0.730);
\draw[black!40,densely dotted] (0.000000,0.510) rectangle (1.000000,0.730);
\fill[tracewait] (1.000000,0.510) rectangle (13.000000,0.730);
\fill[traceblue] (13.000000,0.510) rectangle (14.000000,0.730);
\fill[tracewait] (14.000000,0.510) rectangle (25.083333,0.730);
\draw[black!65,dashed] (13.708333,0.400) -- (13.708333,1.290) node[above,font=\scriptsize] {cancel};
\draw[black!65] (14.000000,0.510) -- (15.500000,0.180) -- (18.000000,0.180);
\node[anchor=west,font=\scriptsize] at (18.000000,0.180) {returns to WAIT};
\node[anchor=north west,align=left,font=\scriptsize,inner sep=0pt] at (27,1.140) {Return delay: 0.29 s\\WAIT through stream end};
\draw[black!40] (0,-.24) -- (25.2,-.24);
\draw[black!40] (0,-.24) -- (0,-.31) node[below,font=\scriptsize] {0};
\draw[black!40] (5,-.24) -- (5,-.31) node[below,font=\scriptsize] {5};
\draw[black!40] (10,-.24) -- (10,-.31) node[below,font=\scriptsize] {10};
\draw[black!40] (15,-.24) -- (15,-.31) node[below,font=\scriptsize] {15};
\draw[black!40] (20,-.24) -- (20,-.31) node[below,font=\scriptsize] {20};
\draw[black!40] (25,-.24) -- (25,-.31) node[below,font=\scriptsize] {25};
\node[anchor=north,font=\scriptsize] at (12.5,-.68) {Video time (s)};
\end{tikzpicture}
\caption{\textbf{Three measured behaviors explain the evaluation components.} Selected diagnostic episodes from the trained reference; model states are held between observations through stream end. (a) Sustained preparation earns lead, while activation during reference \Wait\ incurs a silence cost. (b) Early \Response\ interrupts preparation, giving zero lead despite nonzero response overlap. (c) After cancellation, the model returns to \Wait\ and remains there. Dashed lines mark target time $\tau$ or cancellation; all panels share the same time scale.}
\label{fig:traces}
\end{figure*}

Figure~\ref{fig:traces} connects the evaluation components to three measured behaviors of the trained reference.
In Figure~\ref{fig:traces}(a), sustained preparation earns $7.83$\,s of lead, including $4.92$\,s inside reference \Wait. This additional lead comes with a silence cost.
In (b), early \Response\ interrupts preparation, yielding zero lead while retaining $0.377$ response alignment.

Cancellation-class recall and post-cancellation activity describe complementary aspects of withdrawal.
On all 200 cancellations, these are $0.890$ and $1.33$\,s for the expanded reference, compared with $0.005$ and $5.59$\,s for native video plus LoRA.
The expanded reference prepares before cancellation in 175 episodes and still holds \Prepare\ at the boundary in 100, defining the conditional withdrawal set.
In 96 of those 100 episodes, the model returns to \Wait\ before stream end; in 95 it eventually returns to \Wait\ and remains there through stream end.
The median first-return delay among these 96 returns is $1.79$\,s.
Supplementary tables report conditional withdrawal, all-sample diagnostics, and the policy scores.

In Figure~\ref{fig:traces}(c), the trained reference returns to \Wait\ $0.29$\,s after cancellation and remains there through stream end.
These selected episodes illustrate the diagnostics. Aggregate results are reported in Table~\ref{tab:main}.

\subsection{Scenario Comparisons, Policies, and Sensitivity}

\paragraph{Matched scenarios.}
Sixty-seven pairs share the scenario title, tool, workshop description, clothing, lighting, and background-wall category.
Each pairs a handover with ordinary activity where tools remain visible.
Every ordinary video is used once, and the pairs cover the eight tools with an ordinary track.
Durations are not matched, so we report the two raw behavioral quantities separately in Figure~\ref{fig:matched}.

The matched scenarios distinguish preparation for handover from activation during ordinary activity.
The expanded reference has a mean lead of $2.27$\,s on handovers and is active for $0.15$\,s on ordinary activity.
The native video reference has a mean lead of $2.54$\,s on handovers and is active for $1.87$\,s on ordinary activity.
The expanded and native references activate during ordinary activity in $8/67$ and $37/67$ videos, respectively.
The supplementary material reports descriptive per-pair differences.

\paragraph{Deterministic policies.}
Six content-blind policies use the training-majority tool, a tape measure, whenever active.
Timed responses start at the first observation at or after $9.634$\,s; the clock starts preparation at or after $4.459$\,s.
These are median onsets from the positive source set after test-hash filtering.
Immediate preparation starts at the first observation before that timed response, whereas wait-then-response omits preparation.
No policy receives a test tool, boundary, or event class.

Always preparing scores $0$ ($P=0.882$, with $S=0$ and zero response alignment).
Always responding earns no preparation and scores $0.012$.
Waiting until the fixed response time scores $0.174$, and the clock-based policy scores $0.147$.
Always waiting scores $0.300$ through non-target recognition and silence, with zero target preparation and response alignment.

\paragraph{Sensitivity to $\Lambda$.}
Rescoring stored trajectories at every integer $\Lambda$ from $4$ to $20$\,s changes only preparation.
For the ten models without benchmark-specific training, the trained reference, and JoyAI + task head, the ranking throughout $9$--$20$\,s matches the ranking at $\Lambda=10$\,s, with Kendall's $\tau$ ranging from $0.879$ to $1.000$ over the full sweep.
The trained reference and JoyAI + task head remain first and second across the sweep, and the expanded reference remains ahead of the native reference.

\section{Conclusion}
\label{sec:conclusion}

\bench\ brings sustained, reversible preparation between early evidence and response into the evaluation of streaming tool assistance.
The reference systems show that longer preparation lead can accompany poorer response alignment, and similar handover leads can accompany different levels of activity during ordinary work.
Together with cancellation analyses, these findings place the evolution of preparation (its persistence, transition to response, and withdrawal) at the center of evaluation.

\section*{LLM Usage Statement}
OpenAI Codex was used to refine author-written code, organize author-selected references, and revise author-drafted text. All AI-assisted changes were reviewed and verified by the authors.
\bibliographystyle{plainnat}
\begingroup
\raggedright
\bibliography{main}
\endgroup
\clearpage
% Reuse the supplement while preserving continuous preprint page numbering.
\newcommand{\maketitlesupplementary}{\section*{Supplementary Material}\suppressfloats[t]}
\let\FGIOriginalSetcounter\setcounter
\renewcommand{\setcounter}[2]{\ifstrequal{#1}{page}{}{\FGIOriginalSetcounter{#1}{#2}}}
\setcounter{page}{1}
\maketitlesupplementary
\setcounter{section}{0}
\renewcommand{\thesection}{S\arabic{section}}

\section{Model Inputs and Reference Configuration}

Every learned system receives video frames. Audio is not supplied.
The evaluation clock delivers a causal frame every second, except VideoChat3, which uses a $0.5$\,s period.
The service resets its session between episodes.
Polled systems return task decisions on those observations, while native systems retain their original emission and no-update semantics.
The held-state scorer uses the resulting decisions, not the number of messages emitted.

\paragraph{Reference architectures and observation history.}
The native video reference uses the original JoyAI visual path and decoder, with language attention q/v LoRA of rank 8 and alpha 16.
The expanded video reference keeps that path and additionally encodes recent video and older observed chunks with the frozen V-JEPA~2 ViT-L encoder.
Both use eight recent frame slots at $1$\,fps and $224$-pixel JoyAI inputs, starting at $1$\,s.
Repeated startup slots preserve their actual timestamps.
The added encoder uses $256$-pixel inputs.
The learned bridge supplies 16 recent evidence tokens and four per historical chunk, with at most eight four-frame chunks.
It inserts these features into the native decoder input without using a future-video predictor.

\paragraph{Training data and objective.}
The two short-schedule references sample one causal prefix per training video using the same schedule and seed 20261027.
The training pool and the positive source set used for timing policies are subsets of the tool-video collection. Table~\ref{tab:training-data} gives their sizes and the actual joint-training exposure.
Supervised token loss trains the JSON state/tool answer; teacher-QA answers provide the retention examples.
Visual encoders stay frozen.
The native arm trains decoder LoRA, whereas the expanded arm trains the bridge and decoder LoRA after its own 547-update bridge-alignment stage.
The joint microbatch size is 32, with bridge and LoRA learning rates $10^{-4}$ and $10^{-5}$, respectively, BF16 computation, and a 1,024-token context limit.
The separate trained reference is the expanded architecture after the same alignment and a partial joint checkpoint.
Thus the reported comparisons have explicitly different alignment or joint exposure.

\begin{table}[t]
\centering
\caption{\textbf{Trained video references.} All retain JoyAI's native visual path. Expanded adds V-JEPA~2 recent/history features. Align and Joint are update counts; the two short-schedule references share their joint sampling schedule.}
\label{tab:reference-config}
\scriptsize
\setlength{\tabcolsep}{3pt}
\begin{tabular}{@{}llrr@{}}
\toprule
Reference & Trained parameters & Align & Joint \\
\midrule
Native video + LoRA & Decoder LoRA & 0 & 547 \\
Expanded video & Bridge + decoder LoRA & 547 & 547 \\
Trained reference & Bridge + decoder LoRA & 547 & 2,688 \\
\bottomrule
\end{tabular}
\end{table}

\begin{table}[t]
\centering
\caption{\textbf{Training data and joint exposure.} Video counts refer to the tool-video collection. QA examples are teacher-generated retention examples.}
\label{tab:training-data}
\scriptsize
\setlength{\tabcolsep}{3pt}
\begin{tabular}{@{}lr@{}}
\toprule
Training pool & Videos \\
\midrule
Handover / ordinary / cancellation & 9,873 / 2,469 / 1,660 \\
Total & 14,002 \\
Positive source for timing policies & 9,860 \\
\midrule
Joint schedule & Videos / prefixes / QA \\
\midrule
Short-schedule references & 14,002 / 14,002 / 3,488 \\
Trained reference (partial) & 4,224 / 68,832 / 17,184 \\
\bottomrule
\end{tabular}
\end{table}

\paragraph{Output and decoding.}
The trained references generate a JSON object with state in \{\Wait,\Prepare,\Response\} and a tool ID, with a null tool for \Wait.
Their system instructions define preparation as reversible, handover as response, and subsequent tool use as waiting.
Replay uses deterministic decoding, at most 128 new tokens, a new session for each episode, and an unknown state before the configured first observation.
A fixed ten-entry dictionary maps English tool IDs to the scorer's Chinese names, leaving state and timing unchanged and preserving original text.
JoyAI + task head uses frozen visual and query features with a task-trained mapping and structured prediction heads, trained on tool-assistance trajectories.
At each poll it forms eight causal slots at nominal $4$\,fps, repeating the latest available observation when needed.
Table~\ref{tab:model-interfaces} summarizes history and output handling for the evaluated systems.
Adapters use deterministic rules without access to ground truth.
For LiveCC, an unparseable raw output is mapped to no-update and thus retains the prior state, whereas a prediction that remains invalid after the output adapter interrupts preparation as specified in the evaluation protocol.

\begin{table*}[t]
\centering
\caption{\textbf{Model interfaces.} Polled systems produce a decision on each observation; native systems use their own emission mechanisms. No-update retains the previous state, with an unknown initial state.}
\label{tab:model-interfaces}
\small
\begin{tabular}{@{}p{.19\textwidth}p{.25\textwidth}p{.45\textwidth}@{}}
\toprule
System & History / trigger & State parsing and no-update \\
\midrule
Qwen, LLaVA, Gemma & Observed conversation; polled & Parse task JSON. \\
MiniCPM-o & Streaming KV; polled & Extract task JSON from marked output. \\
JoyAI native & Native streaming; emission & Parse response JSON; silence retains state. \\
MMDuet2 & DTD/KV; two-frame turns & Parse task output; NO REPLY retains state. \\
VideoChat3 & Streaming session; up to 16 rounds & Parse response JSON; silence or standby retains state. \\
LiveCC & Incremental KV; initial six frames, then pairs & Extract task JSON; silence or an unparseable state gives no-update. \\
MOSS-Video & Native frame and token stream & Repair event-class quotes; map \texttt{no\_event} in the state field to \Wait; silence retains state. \\
JoyAI + task head & Causal frame window; polled & Map structured head predictions to state and tool. \\
Trained video references & Recent frames; expanded model adds history chunks; polled & Parse generated state/tool JSON. \\
\bottomrule
\end{tabular}
\end{table*}

\paragraph{Task instructions.}
Instructions ask models to infer the current \Wait, \Prepare, or \Response\ state from observed frames and identify the tool when active. The general-purpose interface requests a JSON state, tool, and event class, while the trained references output state and tool.

\begin{table}[t]
\centering
\caption{\textbf{Deterministic policies on all 1,000 episodes.} Active policies use the training-majority tool. Timed responses start at the training-median onset; the clock also waits for the training-median preparation onset. All follow the same observation schedule and held-state scoring as the video reference.}
\label{tab:policies}
\scriptsize
\setlength{\tabcolsep}{4pt}
\begin{tabular}{@{}lrrrr@{}}
\toprule
Policy & Overall $\uparrow$ & $R\uparrow$ & $P\uparrow$ & $S\uparrow$ \\
\midrule
Always \Wait & 0.300 & 0.200 & 0.000 & 1.000 \\
Always \Prepare & 0.000 & 0.000 & 0.882 & 0.000 \\
Always \Response & 0.012 & 0.023 & 0.000 & 0.000 \\
Immediate \Prepare\ $\to$ timed \Response & 0.019 & 0.038 & 0.324 & 0.000 \\
\Wait\ $\to$ timed \Response & 0.174 & 0.061 & 0.000 & 0.884 \\
Training-median clock & 0.147 & 0.038 & 0.107 & 0.669 \\
\bottomrule
\end{tabular}
\end{table}

\begin{table}[t]
\centering
\caption{\textbf{Preparation, response, and withdrawal.} Lead (s), coverage (Cov.), response alignment ($A$), and premature-response rate (Early) use all 600 targets, including zeros. Trajectory-derived cancellation-class recall (Rec.) and mean active seconds after cancellation (Post) use all 200 cancellations.}
\label{tab:diagnostics}
\scriptsize
\setlength{\tabcolsep}{4pt}
\begin{tabular}{@{}lrrrrrr@{}}
\toprule
System & Lead & Cov. & $A$ & Early & Rec. & Post \\ & $\uparrow$ & $\uparrow$ & $\uparrow$ & $\downarrow$ & $\uparrow$ & $\downarrow$ \\
\midrule
Trained reference & 2.08 & 0.56 & 0.701 & 0.27 & 0.86 & 0.64 \\
Expanded video & 2.08 & 0.56 & 0.695 & 0.16 & 0.89 & 1.33 \\
JoyAI + task head & 5.36 & 0.93 & 0.536 & 0.07 & 0.08 & 5.10 \\
Native video + LoRA & 3.67 & 0.68 & 0.083 & 0.00 & 0.01 & 5.59 \\
Qwen3-VL-8B & 0.04 & 0.06 & 0.239 & 0.50 & 0.00 & 0.70 \\
Gemma-3n & 1.89 & 0.34 & 0.253 & 0.80 & 0.01 & 1.96 \\
\bottomrule
\end{tabular}
\end{table}

\section{Withdrawal After Preparation}

Let $b_i$ be the end of the reference preparation interval of a cancelled episode.
We first report whether any \Prepare\ occurs strictly before $b_i$ (Ever).
The conditional risk set (Held) consists of episodes whose last observation strictly before $b_i$ holds \Prepare.
An observed return is an explicit \Wait\ at or after $b_i$ and strictly before stream end.
An episode that exits before cancellation is outside this risk set; a \Response, unknown, or invalid state does not count as a return.
Stable counts episodes with a return to \Wait\ followed by continuous \Wait\ through stream end; earlier returns may be followed by reactivation.
Delay is the median time from $b_i$ to the first return, computed over observed returns.
Episodes with no observed return remain in the return-rate denominator; delay describes observed exits, not all eligible episodes.
The definitions use the saved observation clock and reference boundaries, without an interpolated decision or a grace period.

\begin{table*}[t]
\centering
\caption{\textbf{Conditional withdrawal for the full model panel.} Ever, Held, Return, Stable, and Delay follow the definitions in the text. Each fraction retains its denominator. A dash indicates no eligible preparation at cancellation, or no observed return for the delay.}
\label{tab:withdrawal-full}
\scriptsize
\setlength{\tabcolsep}{4pt}
\begin{tabular}{@{}lrrrrr@{}}
\toprule
System & Ever & Held & Return & Stable & Median delay (s) \\
\midrule
Qwen3-VL-8B & 77/200 & 14/200 & 12/14 & 8/14 & 1.56 \\
Qwen2.5-Omni-7B & 39/200 & 11/200 & 1/11 & 0/11 & 1.96 \\
LLaVA-OneVision-2-8B-Instruct & 1/200 & 1/200 & 0/1 & 0/1 & -- \\
JoyAI-VL-Interaction & 0/200 & 0/200 & -- & -- & -- \\
MMDuet2 & 0/200 & 0/200 & -- & -- & -- \\
LiveCC-7B-Instruct & 0/200 & 0/200 & -- & -- & -- \\
VideoChat3-4B & 0/200 & 0/200 & -- & -- & -- \\
MOSS-Video & 180/200 & 53/200 & 40/53 & 24/53 & 3.42 \\
Gemma-3n & 200/200 & 48/200 & 17/48 & 13/48 & 1.25 \\
MiniCPM-o-2\_6 & 200/200 & 27/200 & 21/27 & 8/27 & 1.62 \\
Trained reference & 168/200 & 81/200 & 81/81 & 80/81 & 0.88 \\
JoyAI + task head & 187/200 & 157/200 & 24/157 & 6/157 & 2.21 \\
Expanded video & 175/200 & 100/200 & 96/100 & 95/100 & 1.79 \\
Native video + LoRA & 25/200 & 8/200 & 3/8 & 2/8 & 8.46 \\
\bottomrule
\end{tabular}
\end{table*}

\begin{table*}[t]
\centering
\caption{\textbf{Tool identity during preparation for the full panel.} Means and coverage include all 600 targets. Correct-tool lead checks every observation in the trailing run; it does not multiply the state lead by the identity on the first active observation.}
\label{tab:identity-full}
\scriptsize
\setlength{\tabcolsep}{4pt}
\begin{tabular}{@{}lrrr@{}}
\toprule
System & State lead (s) & Correct-tool lead (s) & Correct-tool coverage \\
\midrule
Qwen3-VL-8B & 0.04 & 0.03 & 0.035 \\
Qwen2.5-Omni-7B & 1.19 & 0.21 & 0.023 \\
LLaVA-OneVision-2-8B-Instruct & 0.05 & 0.04 & 0.003 \\
JoyAI-VL-Interaction & 0.00 & 0.00 & 0.000 \\
MMDuet2 & 0.00 & 0.00 & 0.000 \\
LiveCC-7B-Instruct & 0.00 & 0.00 & 0.000 \\
VideoChat3-4B & 0.00 & 0.00 & 0.000 \\
MOSS-Video & 0.52 & 0.03 & 0.023 \\
Gemma-3n & 1.89 & 0.16 & 0.045 \\
MiniCPM-o-2\_6 & 1.16 & 0.00 & 0.000 \\
Trained reference & 2.08 & 1.87 & 0.528 \\
JoyAI + task head & 5.36 & 4.94 & 0.868 \\
Expanded video & 2.08 & 1.84 & 0.530 \\
Native video + LoRA & 3.67 & 0.93 & 0.242 \\
\bottomrule
\end{tabular}
\end{table*}

\begin{table}[t]
\centering
\caption{\textbf{Workplace-matched pairs.} Each of 67 pairs shares the scenario title, tool, workplace description, and clothing, lighting, and background-wall category. Lead is the mean continuous \Prepare-only lead on positives; Ordinary is the mean active time on ordinary videos, in seconds. Durations differ. Diff. is the descriptive mean per-pair difference, not a normalized score or causal effect; Active counts ordinary videos with any activation. Both references receive video.}
\label{tab:pairs}
\small
\setlength{\tabcolsep}{4pt}
\begin{tabular}{@{}lrrrr@{}}
\toprule
System & Lead $\uparrow$ & Ordinary $\downarrow$ & Diff. & Active $\downarrow$ \\
\midrule
Expanded video & 2.27 & 0.15 & 2.12 & 8/67 \\
Native video + LoRA & 2.54 & 1.87 & 0.68 & 37/67 \\
Qwen3-VL-8B-Instruct & 0.08 & 0.63 & -0.55 & 17/67 \\
Gemma-3n-E4B-it & 2.24 & 9.42 & -7.18 & 61/67 \\
\bottomrule
\end{tabular}
\end{table}

\begin{table}[t]
\centering
\caption{\textbf{Preparation and response measure distinct behavior.} The reference response is $[4,8)$\,s. Rows show illustrative held states, with no response outside the stated interval. In $R$, response IoU $A$ is multiplied by event-class and first pre-target active tool correctness.}
\label{tab:examples}
\footnotesize
\begin{tabular}{@{}p{0.65\linewidth}rr@{}}
\toprule
Model trajectory & Lead (s) & $A$ \\
\midrule
\Prepare\ $[1,4)$; \Response\ $[4,8)$ & 3 & 1.00 \\
\Prepare\ $[1,3)$; \Response\ $[3,8)$ & 0 & 0.80 \\
\Response\ $[1,8)$ & 0 & 0.57 \\
\Prepare\ $[1,4)$; no response & 3 & 0.00 \\
\bottomrule
\end{tabular}
\end{table}

The expanded reference has 175 episodes with preparation before cancellation, 100 still preparing at cancellation, 96 observed returns, and 95 stable returns.
Its median observed return delay is $1.79$\,s.
These counts accompany trajectory-derived cancellation-class recall and post-cancellation active seconds over all 200 episodes. The latter alone can be small for an inactive model.
Conditional results retain each system's preparation coverage and follow-up to stream end.

\section{Tool Identity During Sustained Preparation}

The primary v4 preparation lead measures the last uninterrupted \Prepare\ run reaching the target boundary.
Recognition separately checks the tool named on the first active observation before that boundary, together with event class and response-interval overlap.
For an additional diagnostic, we walk backward from the final pre-target observation while both the state is \Prepare\ and the held tool matches the reference tool.
The resulting correct-tool lead is zero if either condition fails at the final pre-target observation.
Its mean and positive-lead coverage use all 600 targets, including zeros.
Tool changes interrupt this run even if the preparation state remains unchanged.
This calculation adds no identity gate to the primary score.

\section{Matched Workplace Comparison}

Each pair shares its scenario title, tool, workshop description, and categories of clothing, lighting, and background wall.
No ordinary episode is reused across pairs.
Video duration is retained: the scorable streams span $11.13$--$26.46$\,s for handovers and $8.04$--$16.54$\,s for ordinary activity.
We report mean preparation lead on handovers and mean active seconds on ordinary episodes as distinct measurements.
Their per-pair difference is descriptive, without duration normalization or a causal interpretation.

\section{Illustrative Preparation and Response Scores}

Early \Response\ earns no preparation lead.
Missing response earns no response-aligned recognition even if the model prepared continuously to the target.
Preparation persistence and tool identity remain separate components of the primary score.

\let\setcounter\FGIOriginalSetcounter
\end{document}